\documentclass[10pt,twocolumn,letterpaper]{article}

\usepackage{cvpr}
\usepackage{times}
\usepackage{epsfig}
\usepackage{graphicx}
\usepackage{amsmath}
\usepackage{amssymb}
\usepackage{algorithm}
\usepackage{algpseudocode}
\usepackage{rotating}
\usepackage[font={small}]{caption}
\usepackage{subcaption}
\usepackage{enumitem}
\usepackage{afterpage}

\usepackage{multirow}
\usepackage{epigraph}
\usepackage{pgfplots}
\usepackage{upgreek}
\usepackage{adjustbox}
\usepackage{gensymb}

\usepackage[breaklinks=true,bookmarks=false]{hyperref}

\cvprfinalcopy 

\def\cvprPaperID{2602} 

\begin{document}

\title{Geometric loss functions for camera pose regression with deep learning}

\author{Alex Kendall ~and~ Roberto Cipolla\\
University of Cambridge\\
{\tt\small \{agk34,~rc10001\}@cam.ac.uk}
}

\newcommand{\fig}[1]{Figure~\ref{fig:#1}}
\newcommand{\tbl}[1]{Table~\ref{tbl:#1}}
\newcommand{\sct}[1]{Section~\ref{sec:#1}}
\newcommand{\eqn}[1]{(\ref{eqn:#1})}

\maketitle

\begin{abstract}

Deep learning has shown to be effective for robust and real-time monocular image relocalisation. In particular, PoseNet \cite{kendall2015posenet} is a deep convolutional neural network which learns to regress the 6-DOF camera pose from a single image. It learns to localize using high level features and is robust to difficult lighting, motion blur and unknown camera intrinsics, where point based SIFT registration fails. However, it was trained using a naive loss function, with hyper-parameters which require expensive tuning. In this paper, we give the problem a more fundamental theoretical treatment. We explore a number of novel loss functions for learning camera pose which are based on geometry and scene reprojection error. Additionally we show how to automatically learn an optimal weighting to simultaneously regress position and orientation. By leveraging geometry, we demonstrate that our technique significantly improves PoseNet's performance across datasets ranging from indoor rooms to a small city.

\end{abstract}

\section{Introduction}
\label{sec:introduction}

\begin{figure}[t]
	\begin{center}
		\includegraphics[width=0.8\linewidth]{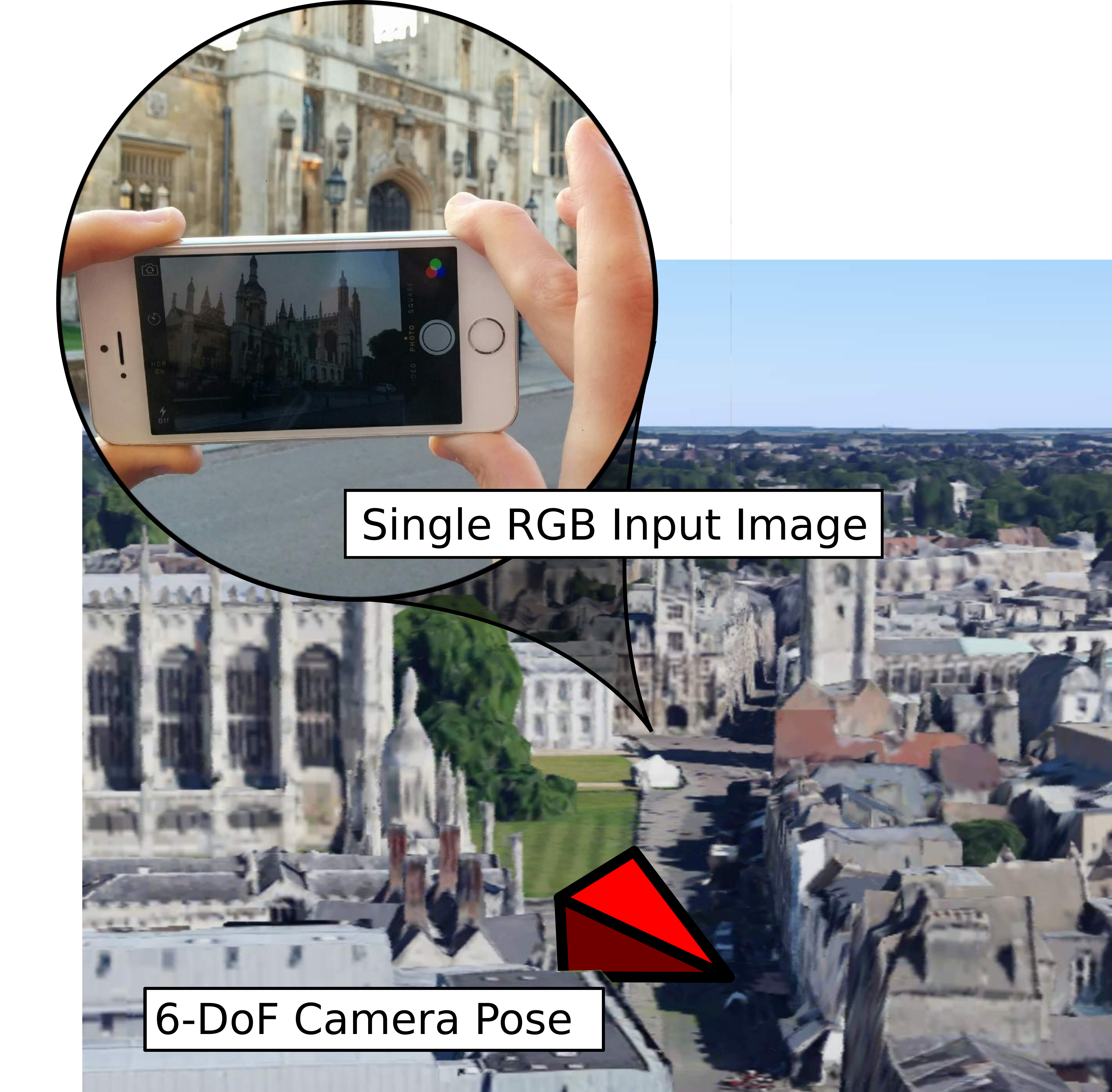}
	\end{center}
    \vspace{-4mm}
	\caption{\textbf{PoseNet} \cite{kendall2015posenet} is trained end-to-end to estimate the camera's six degree of freedom pose from a single monocular image. In this paper we show how to apply a principled loss function based on the scene's geometry to learn camera pose without any hyper-parameters.}
    \vspace{-5mm}
	\label{fig:teaser}
\end{figure}

Designing a system for reliable large scale localisation is a challenging problem. The discovery of the positioning system in mammalian brains, located in the hippocampus, was awarded the 2014 Nobel prize in Physiology or Medicine \cite{o1978hippocampus,moser2008place}. It is an important problem for computer vision too, with localisation technology essential for many applications including autonomous vehicles, unmanned aerial vehicles and augmented reality. State of the art localisation systems perform very well within controlled environments \cite{klein2007parallel,newcombe2011dtam,engel2014lsd,mur2015orb,Sattler14ECCV}. However, we are yet to see their wide spread use in the wild because of their inability to cope with large viewpoint or appearance changes.

Many of the visual localisation systems use point landmarks such as SIFT \cite{lowe2004distinctive} or ORB \cite{rublee2011orb} to localise. These features perform well for incremental tracking and estimating ego-motion \cite{mur2015orb}. However, these point features are not able to create a representation which is sufficiently robust to challenging real-world scenarios. For example, point features are often not robust enough for localising across different weather, lighting or environmental conditions. Additionally, they lack the ability to capture global context, and require robust aggregation of hundreds of points to form a consensus to predict pose \cite{zeisl2015camera}.

To address this problem, we introduced PoseNet \cite{kendall2015posenet,kendall2015modelling} which uses end-to-end deep learning to predict camera pose from a single input image. It was shown to be able localise more robustly using deep learning, compared with point features such as SIFT \cite{lowe2004distinctive}. PoseNet learns a representation using the entire image context based on appearance and shape. These features generalise well and can localise across challenging lighting and appearances changes. It is also fast, being able to regress the camera's pose in only a few milliseconds. It is very scalable as it does not require a large database of landmarks. Rather, it learns a mapping from pixels to a high dimensional space linear with pose.

The main weakness of PoseNet \cite{kendall2015posenet} was that despite its scalability and robustness it did not produce metric accuracy which is comparable to other geometric methods \cite{Sattler14ECCV,Svarm14CVPR}. In this paper we argue that a contributing factor to this was because PoseNet naively applied a deep learning model end-to-end to learn camera pose. In this work, we reconsider this problem with a grounding in geometry. We wish to build upon the decades of research into multi-view geometry \cite{hartley2003multiple} to improve our ability to use deep learning to regress camera pose.

The main contribution of this paper is improving the performance of PoseNet with geometrically formed loss functions. It is not trivial to simply regress position and rotation quantities using supervised learning. PoseNet required a weighting factor to balance these two properties, but was not tolerant to the selection of this hyperparameter. In \sct{loss} we explore loss functions which remove this hyperparameter, or optimise it directly from the data. In \sct{reproj} we show how to train directly from the scene geometry using the reprojection error.

In \sct{exp} we demonstrate our system on an array of datasets, ranging from individual indoor rooms, to the Dubrovnik city dataset \cite{li2012worldwide}. We show that our geometric approach can improve PoseNet's efficacy across many different datasets -- narrowing the deficit to traditional SIFT feature-based algorithms. For outdoor scenes ranging from $50,000m^2$ to $2km^2$ we can achieve relocalisation accuracies of a few meters and a few degrees. In small rooms we are able to achieve accuracies of $0.2-0.4m$.

\section{Related work}

Large scale localisation research can be divided into two categories; place recognition and metric localisation. Place recognition discretises the world into a number of landmarks and attempts to identify which place is visible in a given image. Traditionally, this has been modelled as an image retrieval problem \cite{ChenBaatz11CVPR,cummins2008fab,Torii13CVPR,Schindler07CVPR} enabling the use of efficient and scalable retrieval approaches \cite{Nister06CVPR,Philbin07CVPR} such as Bag-of-Words (BoW)~\cite{Sivic03ICCV}, VLAD~\cite{Jegou-CVPR10,Delhumeau-ACMMM13}, and Fisher vectors~\cite{Jegou-PAMI12}. Deep learning models have also been shown to be effective for creating efficient descriptors. Many approaches leverage classification networks~\cite{RSMC14,GWGL14,BL15,Tolias16ICLR}, and fine tune them on localisation datasets \cite{BSCL14}. Other work of note is PlaNet \cite{weyand2016planet} which trained a classification network to localise images on a world scale. However, all these networks must discretise the world into places and are unable to produce a fine grained estimate of 6-DOF pose.

In contrast, metric localisation techniques estimate the metric position and orientation of the camera. Traditionally, this has been approached by computing the pose from 2D-3D correspondences between 2D features in the query image and 3D points in the model, which are determined through descriptor matching \cite{Choudhary12ECCV,Li10ECCV,Li12ECCV,Sattler12ECCV,Svarm14CVPR}. This assumes that the scene is represented by a 3D structure-from-motion model. The full 6 degree-of-freedom pose of a query image can be estimated very precisely \cite{Sattler14ECCV}. However these methods require a 3D model with a large database of features and efficient retrieval methods. They are expensive to compute, often do not scale well, and are often not robust to changing environmental conditions \cite{walch2016image}.

In this work, we address the more challenging problem of metric localisation with deep learning. PoseNet \cite{kendall2015posenet} introduced the technique of training a convolutional neural network to regress camera pose. It combines the strengths of place recognition and localisation approaches: it can globally relocalise without a good initial pose estimate, and produces a continuous metric pose. Rather than building a map (or database of landmark features), the neural network learns features whose size, unlike a map, does not require memory linearly proportional to the size of the scene.

Later work has extended PoseNet to use RGB-D input \cite{li2017indoor}, learn relative ego-motion \cite{melekhov2017relative}, improve the context of features \cite{walch2016image}, localise over video sequences \cite{clark2017vidloc} and interpret relocalisation uncertainty with Bayesian Neural Networks \cite{kendall2015modelling}. Additionally, \cite{walch2016image} demonstrate PoseNet's efficacy on featureless indoor environments, where they demonstrate that SIFT based structure from motion techniques fail in the same environment.

Although PoseNet is scalable and robust \cite{kendall2015posenet}, it does not produce sufficiently accurate estimates of Pose compared to traditional methods \cite{Sattler14ECCV}. It was designed with a naive regression loss function which trains the network end-to-end without any consideration for geometry. This problem is the focus of this paper -- we do not want to throw away the decades of research into multi view geometry \cite{hartley2003multiple}. We improve PoseNet's performance by learning camera pose with a fundamental treatment of scene geometry.

\section{Model for camera pose regression}
\label{sec:model}

In this section we describe the details of the convolutional neural network model we train to estimate camera pose directly from a monocular image, $I$. Our network outputs an estimate, $\mathbf{\hat{p}}$, for pose, $\mathbf{p}$, given by a 3-D camera position $\mathbf{\hat{x}}$ and orientation $\mathbf{\hat{q}}$. We use a quaternion to represent orientation, for reasons discussed in \sct{rot}. Pose $\mathbf{p}$ is defined relative to an arbitrary global reference frame. In practice we centre this global reference frame at the mean location of all camera poses. We train the model with supervised learning using pose labels, $\mathbf{p} = [\mathbf{x}, \mathbf{q}]$, obtained through structure from motion, or otherwise (\sct{data}). 

\subsection{Architecture}
\label{sec:arch}

Our pose regression formulation is capable of being applied to any neural network trained through back propagation. For the experiments in this paper we adapt a state of the art deep neural network architecture for classification, GoogLeNet \cite{szegedy2014going}, as a basis for developing our pose regression network. This allows us to use pretrained weights, for example from a model trained to classify images in the ImageNet dataset \cite{deng2009imagenet}. We observe that these pretrained features regularise and improve performance in PoseNet through transfer learning \cite{oquab2014learning}. Although, to generalise PoseNet, we may apply it to any deep architecture designed for image classification as follows:

\begin{enumerate}[topsep=0.5pt,itemsep=0.5ex,partopsep=0ex,parsep=0ex]
	\item Remove the final linear regression and softmax layers used for classification
	\item Append a linear regression layer. This fully connected layer is designed to output a seven dimensional pose vector representing position (3 dimensions) and orientation (4 dimensional quaternion)
	\item Insert a normalisation layer to normalise the four dimensional quaternion orientation vector to unit length
\end{enumerate}

\subsection{Pose representation}
\label{sec:rot}

An important consideration when designing a machine learning system is the representation space of the output. We can easily learn camera position in Euclidean space \cite{kendall2015posenet}. However, learning orientation is more complex. In this section we compare a number of different parametrisations used to express rotational quantities; Euler angles, axis-angle, $SO(3)$ rotation matrices and quaternions \cite{altmann2005rotations}. We evaluate their efficacy for deep learning.

Firstly, Euler angles are easily understandable as an interpretable parametrisation of 3-D rotation. However, they have two problems. Euler angles wrap around at $2\pi$ radians, having multiple values representing the same angle. Therefore they are not injective, which causes them to be challenging to learn as a uni-modal scalar regression task. Additionally, they do not provide a unique parametrisation for a given angle and suffer from the well-studied problem of gimbal lock \cite{altmann2005rotations}. The axis-angle representation is another three dimensional vector representation. However like Euler angles, it too suffers from a repetition around the $2\pi$ radians representation.

Rotation matrices are a over-parametrised representation of rotation. For 3-D problems, the set of rotation matrices are $3\times3$ dimensional members of the special orthogonal Lie group, $SO(3)$. These matrices have a number of interesting properties, including orthonormality. However, it is difficult to enforce the orthogonality constraint when learning a $SO(3)$ representation through back-propagation.

In this work, we chose quaternions as our orientation representation. Quaternions are favourable because arbitrary four dimensional values are easily mapped to legitimate rotations by normalizing them to unit length. This is a simpler process than the orthonormalization required of rotation matrices. Quaternions are a continuous and smooth representation of rotation. They lie on the unit manifold, which is a simple constraint to enforce through back-propagation. Their main downside is that they have two mappings for each rotation, one on each hemisphere. However, in \sct{quat} we show how to adjust the loss function to compensate for this.

\subsection{Loss function}
\label{sec:loss}

This far, we have described the structure of the pose representation we would like our network to learn. Next, we discuss how to design an effective loss function to learn to estimate the camera's 6 degree of freedom pose. This is a particularly challenging objective because it involves learning two distinct quantities - rotation and translation - with different units and scales.

This section defines a number of loss functions and explores their efficacy for camera pose regression. We begin in \sct{loss_weighted} by describing the original weighted loss function which was proposed by PoseNet \cite{kendall2015posenet}. We improve on this in \sct{learn_loss} by introducing a novel loss function which can learn the weighting between rotation and translation automatically, using an estimate of the \textit{homoscedastic} task uncertainty. Further, in \sct{reproj} we describe a loss function which combines position and orientation as a single scalar using the reprojection error geometry. In \sct{loss_exp} we compare the performance of these loss functions, and discusses their trade-offs.

\subsubsection{Learning position and orientation}
\label{sec:quat}

We can learn to estimate camera position by forming a smooth, continuous and injective regression loss in Euclidean space, $\mathcal{L}_x(I) = \left\lVert\mathbf{x} - \mathbf{\hat{x}}\right\rVert_\gamma$, with norm given by $\gamma$ (\cite{kendall2015posenet} used the $L_2$ Euclidean norm).

However, learning camera orientation is not as simple. In \sct{rot} we described a number of options for representing orientation. Quaternions are an attractive choice for deep learning because they are easily formulated in a continuous and differentiable way. The set of rotations lives on the unit sphere in quaternion space. We can easily map any four dimensional vector to a valid quaternion rotation by normalising it to unit length. \cite{kendall2015posenet} demonstrates how to learn to regress quaternion values:
\begin{equation}
\mathcal{L}_q(I) = \left\lVert \mathbf{q}-\frac{\mathbf{\hat{q}}}{\left\lVert\mathbf{\hat{q}}\right\rVert}\right\rVert_\gamma
\label{eqn:loss_quaternion_posenet}
\end{equation}
Using a distance norm, $\gamma$, in Euclidean space makes no effort to keep $\mathbf{q}$ on the unit sphere. We find, however, that during training, $\mathbf{q}$ becomes close enough to $\mathbf{\hat{q}}$ such that the distinction between spherical distance and Euclidean distance becomes insignificant. For simplicity, and to avoid hampering the optimization with unnecessary constraints, we chose to omit the spherical constraint. The main problem with Quaternions is that they are not injective because they have two unique values (from each hemisphere) which map to a single rotation. This is because quaternion, $\textbf{q}$, is identical to $-\textbf{q}$. To address this, we constrain all quaternions to one hemisphere such that there is a unique value for each rotation.

\subsubsection{Simultaneously learning position and orientation}
\label{sec:loss_weighted}

The challenging aspect of learning camera pose is designing a loss function which is able to learn both position and orientation. Initially, we proposed a method to combine position and orientation into a single loss function with a linear weighted sum \cite{kendall2015posenet}, shown in \eqn{loss1}:
\begin{equation}
\mathcal{L}_{\beta}(I) = \mathcal{L}_x(I) + \beta \mathcal{L}_q(I)
\label{eqn:loss1}
\end{equation}
Because $\mathbf{x}$ and $\mathbf{q}$ are expressed in different units, a scaling factor, $\beta$, is used to balance the losses. This hyper-parameter attempts to keep the expected value of position and orientation errors approximately equal.

\begin{figure}[t]
\centering
\resizebox{0.8\columnwidth}{!}{
\begin{tikzpicture}
\pgfplotsset{
    compat=1.3,
    scale only axis,
    height=5cm,
    width=\columnwidth 
}

\begin{axis}[
  xmode=log,
  log ticks with fixed point,
  axis y line*=left,
  ymin=1.4, ymax=3.0,
  xlabel=Beta Weight $\beta$,
  ylabel=Median Position Error (m),
  xmin=100, xmax=2000,
    log basis x={2},
    xticklabel=\pgfmathparse{2^\tick}\pgfmathprintnumber{\pgfmathresult},
]
\addplot[mark=x,red,very thick]
  coordinates{
    (100,1.81)
    (250,1.61)
    (500,1.52)
    (750,1.63)
    (1000,1.91)
    (2000,2.92)
}; \label{plot_one}
\addlegendentry{Position}
\end{axis}

\begin{axis}[
  xmode=log,
  log ticks with fixed point,
  axis y line*=right,
  xtick=\empty,
  ymin=1.0, ymax=7.0,
  ylabel=Median Orientation Error ($\degree$),
  xmin=100, xmax=2000,
]
\addlegendimage{/pgfplots/refstyle=plot_one}\addlegendentry{Position}
\addplot[mark=*,blue,very thick]
  coordinates{
    (100,7.32)
    (250,3.43)
    (500,1.19)
    (750,1.20)
    (1000,1.53)
    (2000,2.27)
}; \label{plot_two}
\addlegendentry{Orientation}
\end{axis}
\end{tikzpicture}}
 
	\caption{Relative performance of position and orientation regression on \textbf{a single model with a range of scale factors} for an indoor scene from the King's College scene in Cambridge Landmarks, using the loss function in \eqn{loss1}. This demonstrates that learning with the optimum scale factor leads to the network uncovering a more accurate pose function.}
\label{fig:scalefactor}
\end{figure}
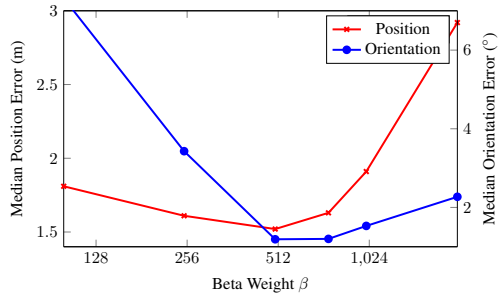

Interestingly, we observe that a model which is jointly trained to regress the camera's position and orientation performs better than separate models trained on each task individually. \fig{scalefactor} shows that with just position, or just orientation information, the network was not able to determine the function representing camera pose with as great accuracy. The model learns a better representation for pose when supervised with both translation and orientation labels. We also experimented with branching the network lower down into two separate components to regress position and orientation. However, we found that it too was less effective, for similar reasons: separating into distinct position and orientation features denies each the information necessary to factor out orientation from position, or vice versa.

However the consequence of this was that the hyper-parameter $\beta$ required significant tuning to get reasonable results. In the loss function \eqn{loss1} a balance $\beta$ must be struck between the orientation and translation penalties (\fig{scalefactor}). They are highly coupled as they are regressed from the same model weights. We found $\beta$ to be greater for outdoor scenes as position errors tended to be relatively greater. Following this intuition it is possible to fine tune $\beta$ using grid search. For the indoor scenes it was between 120 to 750 and outdoor scenes between 250 to 2000. This is an expensive task in practice, as each experiment can take days to complete. It is desirable to find a loss function which removes this hyperparameter. Therefore, the remainder of this section explores different loss functions which aim to find an optimal weighting automatically.

\subsubsection{Learning an optimal weighting}
\label{sec:learn_loss}

Ideally, we would like a loss function which is able to learn position and orientation optimally, without including any hyper parameters. For this reason, we propose a novel loss function which is able to learn a weighting between the position and orientation objective functions. We formulate it using \textit{homoscedastic uncertainty} which we can learn using probabilistic deep learning \cite{kendall2017uncertainties}. Homoscedastic uncertainty is a measure of uncertainty which does not depend on the input data, as opposed to heteroscedastic uncertainty which is a function of the input data \cite{kendall2017uncertainties}. Rather, it captures the uncertainty of the task itself. In \cite{kendall2017multi} we show how to use this insight to combine losses for different tasks in a probabilistic manner. Here we show how to apply this to learn camera position and orientation (with a Laplace likelihood):
\begin{equation}
\label{eqn:loss4}
\mathcal{L}_{\sigma}(I) = \mathcal{L}_x(I) \hat{\sigma}_x^{-2} + \log{\hat{\sigma}_x^2} + \mathcal{L}_{q}(I) \hat{\sigma}_q^{-2} + \log{\hat{\sigma}_q^2}
\end{equation}
where we optimise the homoscedastic uncertainties, $\hat{\sigma}_x^2$, $\hat{\sigma}_q^2$, through backpropagation with respect to the loss function. These uncertainties are free scalar values, not model outputs. They represent the homoscedastic (task) noise.

This loss consists of two components; the residual regressions and the uncertainty regularization terms. We learn the variance, $\sigma^2$, implicitly from the loss function. As the variance is larger, it has a tempering effect on the residual regression term; larger variances (or uncertainty) results in a smaller residual loss. The second regularization term prevents the network from predicting infinite uncertainty (and therefore zero loss). As we expect quaternion values to have much smaller values (they are constrained to the unit manifold), their noise, $\sigma_q^2$ should be much smaller than the position noise, $\sigma_x^2$, which can be many meters in magnitude. As $\sigma_q^2$ should be much smaller than $\sigma_x^2$, orientation regression should be weighted much higher than position -- with a similar effect to $\beta$ in \eqn{loss1}.

In practice, we learn $\hat{s}:=\log\hat{\sigma}^2$ because it is more numerically stable \cite{kendall2017multi}:
\begin{equation}
\label{eqn:loss5}
\mathcal{L}_{\sigma}(I) = \mathcal{L}_x(I) \exp (-\hat{s}_x) + \hat{s}_x + \mathcal{L}_{q}(I) \exp (-\hat{s}_q) + \hat{s}_q
\end{equation}
This is more numerically stable than regressing the variance, $\sigma^2$, because the loss avoids a potential division by zero. The exponential mapping also allows us to regress unconstrained scalar values, where $\exp(-s_i)$ is resolved to the positive domain giving valid values for variance. We find that this loss is very robust to our initialisation choice for the homoscedastic task uncertainty values. Only an approximate initial guess is required, we arbitrarily use initial values of $\hat{s}_x=0.0, \hat{s}_q=-3.0$, for all scenes.

\subsubsection{Learning from geometric reprojection error}
\label{sec:reproj}

Perhaps a more desirable loss is one that does not require balancing of rotational and positional quantities at all. Reprojection error of scene geometry is a representation which combines rotation and translation naturally in a single scalar loss \cite{hartley2003multiple}. Reprojection error is given by the residual between 3-D points in the scene projected onto a 2-D image plane using the ground truth and predicted camera pose. It therefore converts rotation and translation quantities into image coordinates. This naturally weights translation and rotation quantities depending on the scene and camera geometry.

To formulate this loss, we first define a function, $\uppi$, which maps a 3-D point, $\mathbf{g}$, to 2-D image coordinates, $(u,v)^T$:
\begin{equation}
\uppi (\mathbf{x},\mathbf{q},\mathbf{g}) \mapsto \begin{pmatrix} u \\ v \end{pmatrix}
\end{equation}
where $\mathbf{x}$ and $\mathbf{q}$ represent the camera position and orientation. This function, $\uppi$, is defined as:
\begin{equation}
\begin{pmatrix} u' \\ v' \\ w' \end{pmatrix} = \mathsf{K} ( \mathsf{R} \mathbf{g} + \mathbf{x}), 
~~\begin{pmatrix} u \\ v \end{pmatrix} = \begin{pmatrix} u'/w' \\ v'/w' \end{pmatrix}
\end{equation}
where $\mathsf{K}$ is the intrinsic calibration matrix of the camera, and $\mathsf{R}$ is the mapping of $\mathbf{q}$ to its $SO(3)$ rotation matrix, $\textbf{q}_{4\times1} \mapsto \mathsf{R}_{3\times3}$.

We formulate this loss by taking the norm of the reprojection error between the predicted and ground truth camera pose. We take the subset, $\mathcal{G'}$, of all 3-D points in the scene, $\mathcal{G}$, which are visible in the image $I$. The final loss \eqn{loss_reproject} is given by the mean of all the residuals from points, $g_i \in \mathcal{G'}$:
\begin{equation}
\mathcal{L}_g(I) = \dfrac{1}{|\mathcal{G}'|} \sum\limits_{g_i \in \mathcal{G}'} \left\lVert \uppi (\mathbf{x},\mathbf{q},\mathbf{g_i}) - \uppi (\mathbf{\hat{x}},\mathbf{\hat{q}},\mathbf{g_i}) \right\rVert_\gamma
\label{eqn:loss_reproject}
\end{equation}
where $\mathbf{\hat{x}}$ and $\mathbf{\hat{q}}$ are the predicted camera poses from PoseNet, with $\mathbf{x}$ and $\mathbf{q}$ the ground truth label, with norm, $\gamma$, which is discussed in \sct{norm}.

Note that because we are projecting 3-D points using both the ground truth and predicted camera pose we can apply any arbitrary camera model, as long as we use the same intrinsic parameters for both cameras. Therefore for simplicity, we set the camera intrinsics, $K$, to the identity matrix -- camera calibration is not required.

This loss implicitly combines rotation and translational quantities into image coordinates. Minimising reprojection error is often the most desirable balance between these quantities for many applications, such as augmented reality. The key advantage of this loss is that it allows the model to vary the weighting between position and orientation, depending on the specific geometry in the training image. For example, training images with geometry which is far away would balance rotational and translational loss differently to images with geometry very close to the camera.

Interestingly, when experimenting with the original weighted loss in \eqn{loss1} we observed that the hyperparameter $\beta$ was an approximate function of the scene geometry. We observed that it was a function of the landmark distance and size in the scene. Our intuition was that the optimal choice for $\beta$ was approximating the reprojection error in the scene geometry. For example, if the scene is very far away, then rotation is more significant than translation and vice versa. This function is not trivial to model for complex scenes with a large number of landmarks. It will vary significantly with each training example in the dataset. By learning with reprojection error we can use our knowledge of the scene geometry more directly to automatically infer this weighting.

Projecting geometry through a projection model is a differentiable operation involving matrix multiplication. Therefore we can use this loss to train our model with stochastic gradient descent. It is important to note that we do not need to know the intrinsic camera parameters to project this 3-D geometry. This is because we apply the same projection to both the model prediction and ground truth measurement, so we can use arbitrary values.

It should be noted that we need to have some knowledge of the scene's geometry in order to have 3-D points to reproject. The geometry is often known; if our data is obtained through structure from motion, RGBD data or other sensory data (see \sct{data}). Only points from the scene which are visible in the image $I$ are used to compute the loss. We also found it was important for numerical stability to ignore points which are projected outside the image bounds.

\begin{figure*}[t]
\centering

	\begin{subfigure}[t]{\linewidth}
	\resizebox{\linewidth}{!}{
		\includegraphics[width=0.2\linewidth]{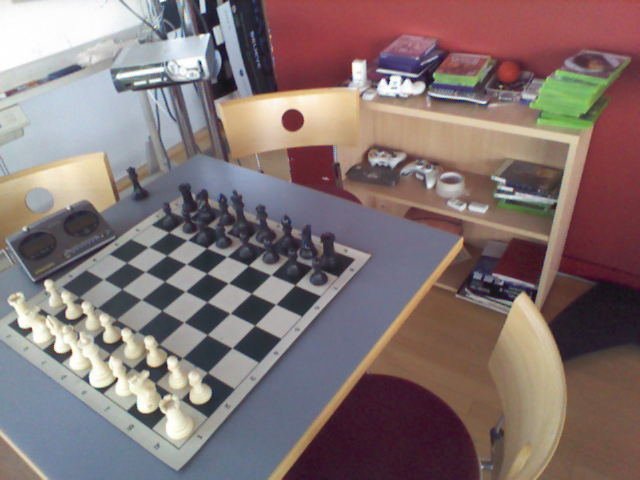}
		\includegraphics[width=0.2\linewidth]{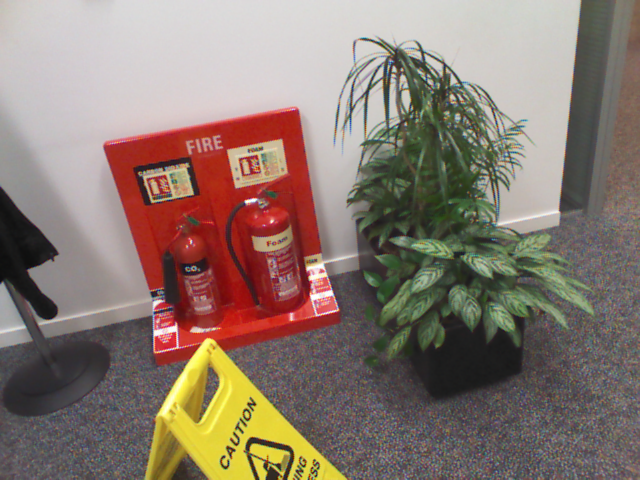}
		\includegraphics[width=0.2\linewidth]{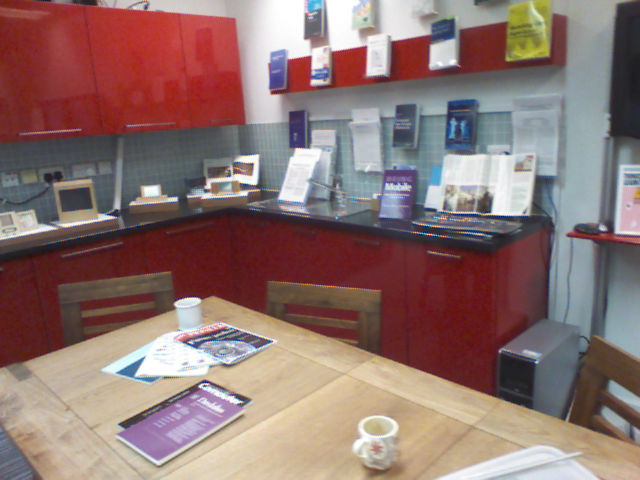}
		\includegraphics[width=0.2\linewidth]{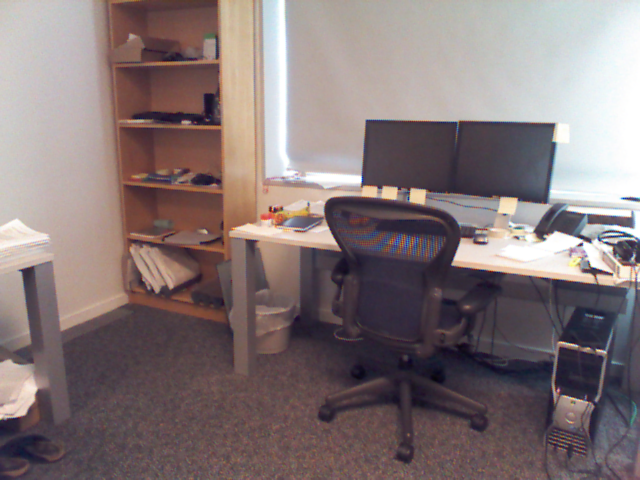}
		\includegraphics[width=0.2\linewidth]{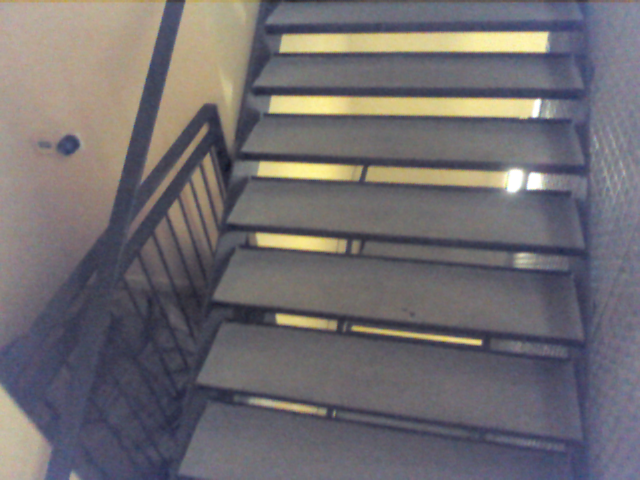}
		}
	\caption{7 Scenes Dataset - 43,000 images from seven scenes in small indoor environments \cite{shotton2013scene}.}
	~
	~
	\end{subfigure}
	
	\begin{subfigure}[t]{\linewidth}
	\resizebox{\linewidth}{!}{
		\includegraphics[width=0.2\linewidth]{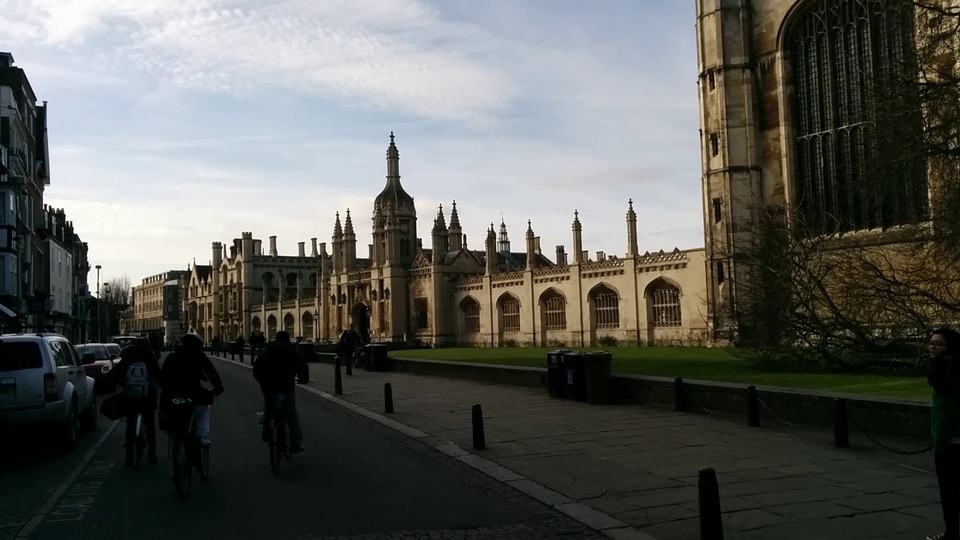}
		\includegraphics[width=0.2\linewidth]{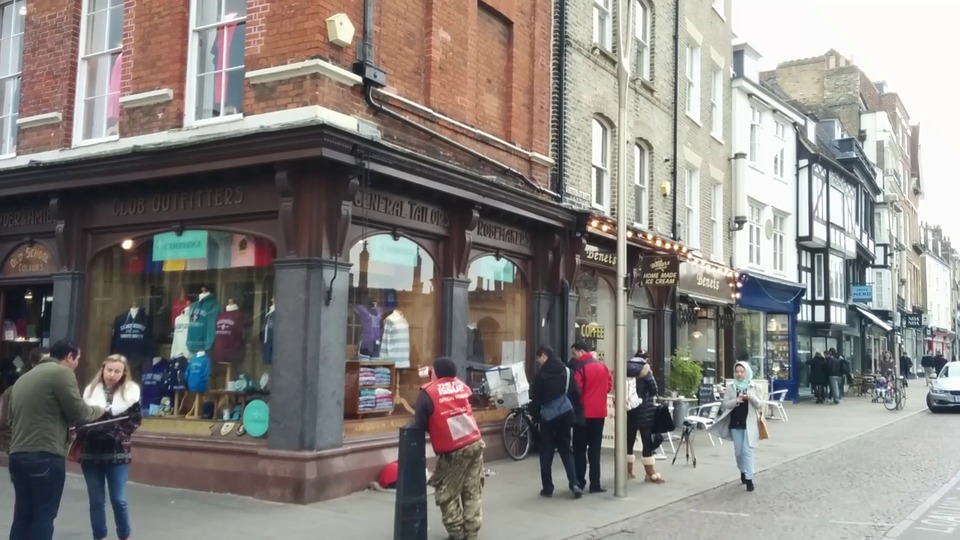}
		\includegraphics[width=0.2\linewidth]{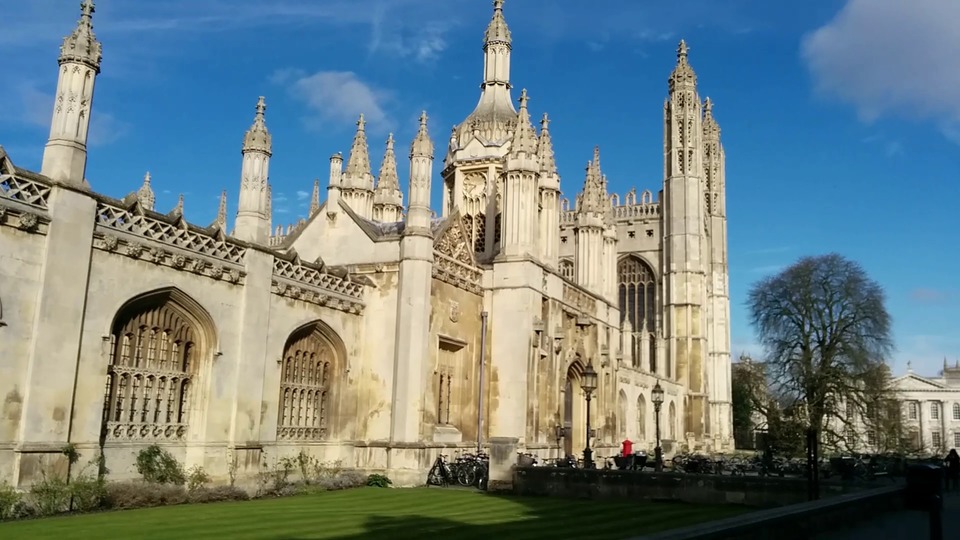}
		\includegraphics[width=0.2\linewidth]{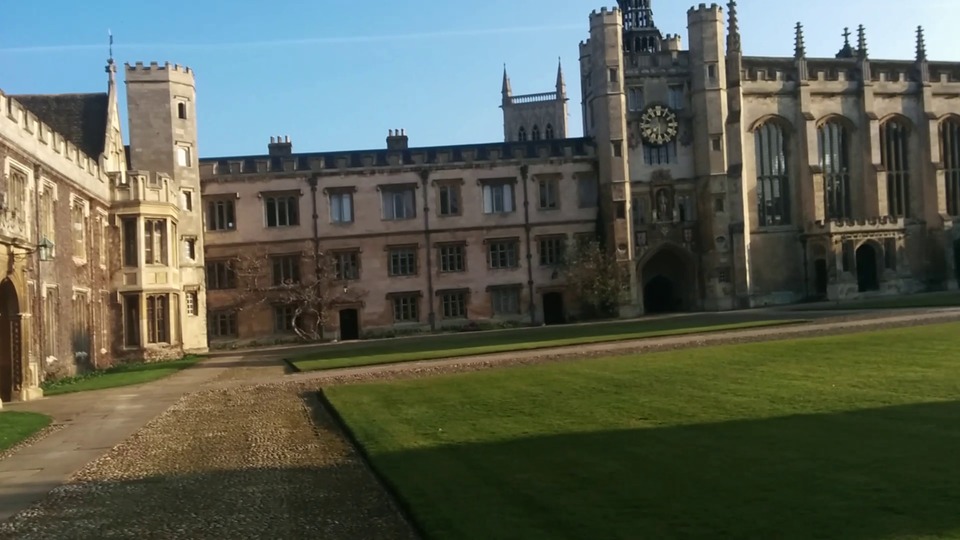}
		\includegraphics[width=0.2\linewidth]{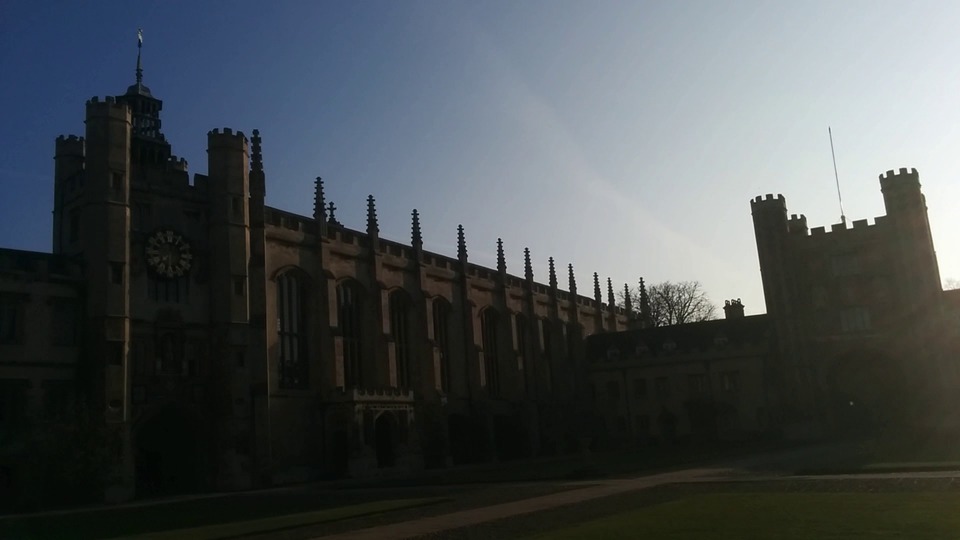}
		}
        
    \vspace{0.5mm}
	\resizebox{\linewidth}{!}{
		\includegraphics[width=0.2\linewidth]{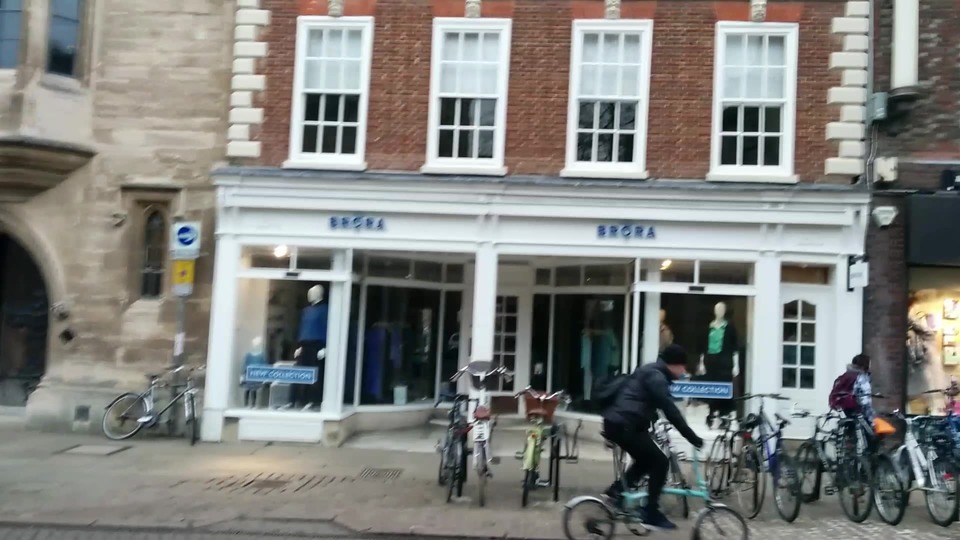}
		\includegraphics[width=0.2\linewidth]{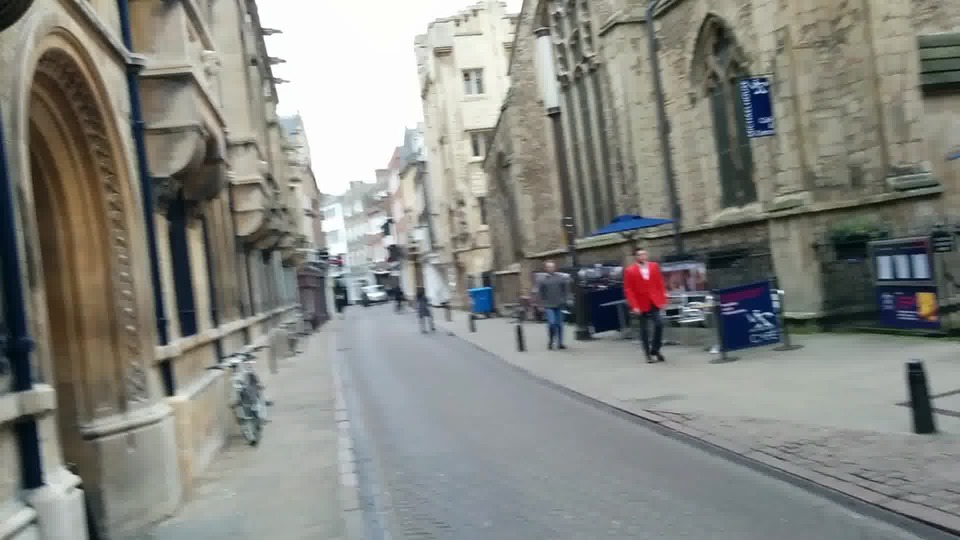}
		\includegraphics[width=0.2\linewidth]{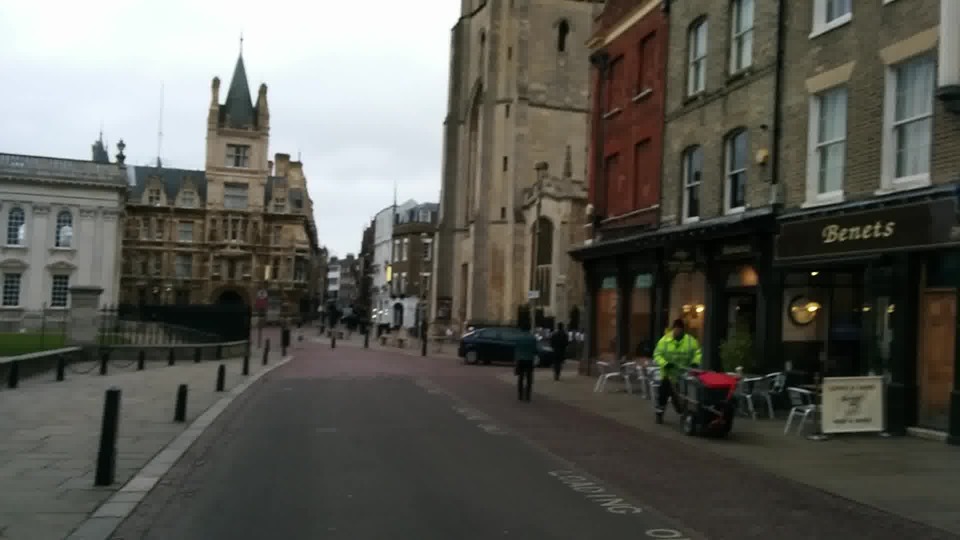}
		\includegraphics[width=0.2\linewidth]{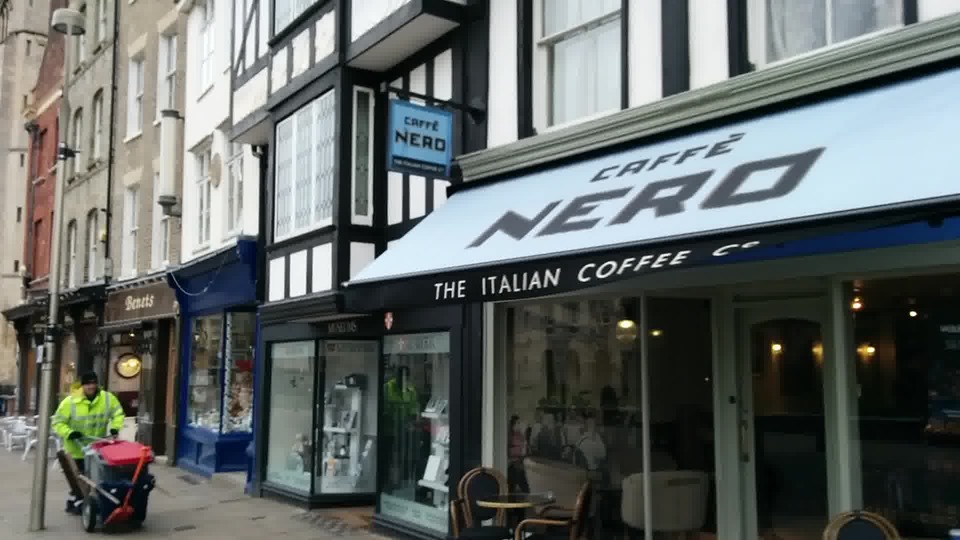}
		\includegraphics[width=0.2\linewidth]{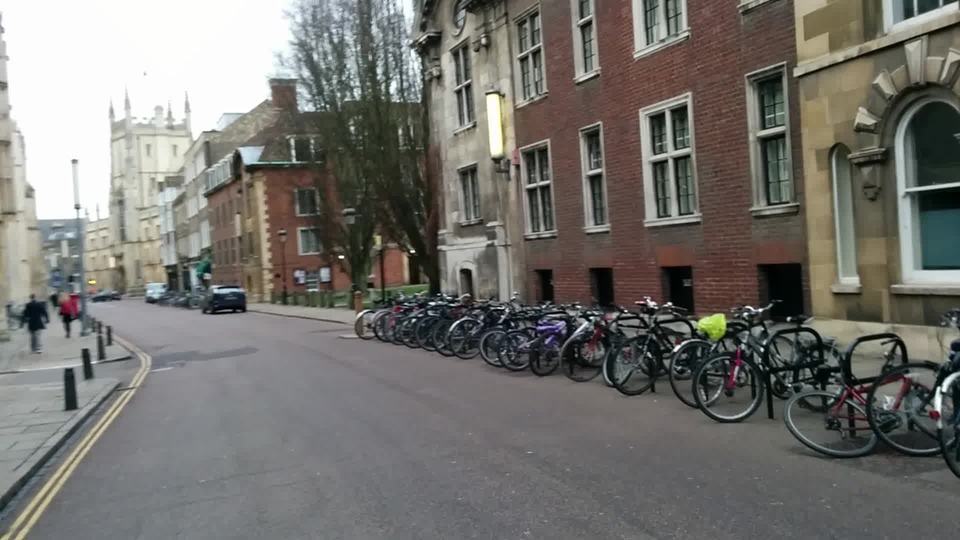}
		}
	\caption{Cambridge Landmarks Dataset - over 10,000 images from six scenes around Cambridge, UK \cite{kendall2015posenet}.}
	~
	~
	\end{subfigure}
	
	\begin{subfigure}[t]{\linewidth}
	\resizebox{\linewidth}{!}{
		\includegraphics[height=50mm]{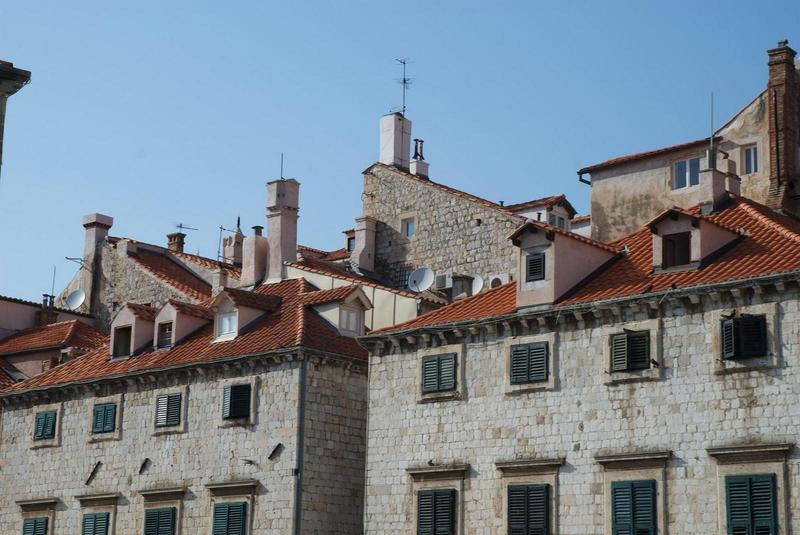}
		\includegraphics[height=50mm]{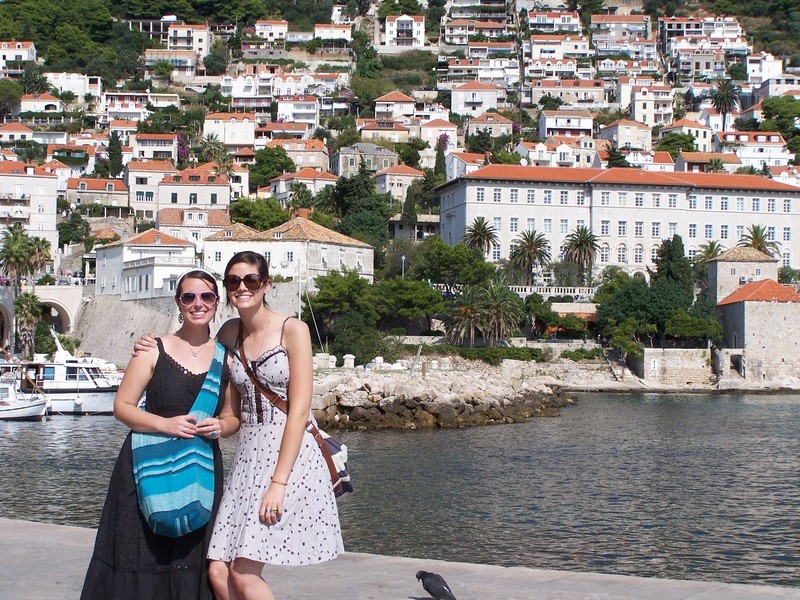}
		\includegraphics[height=50mm]{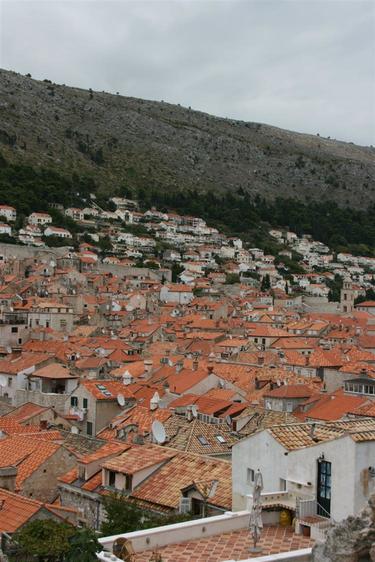}
		\includegraphics[height=50mm]{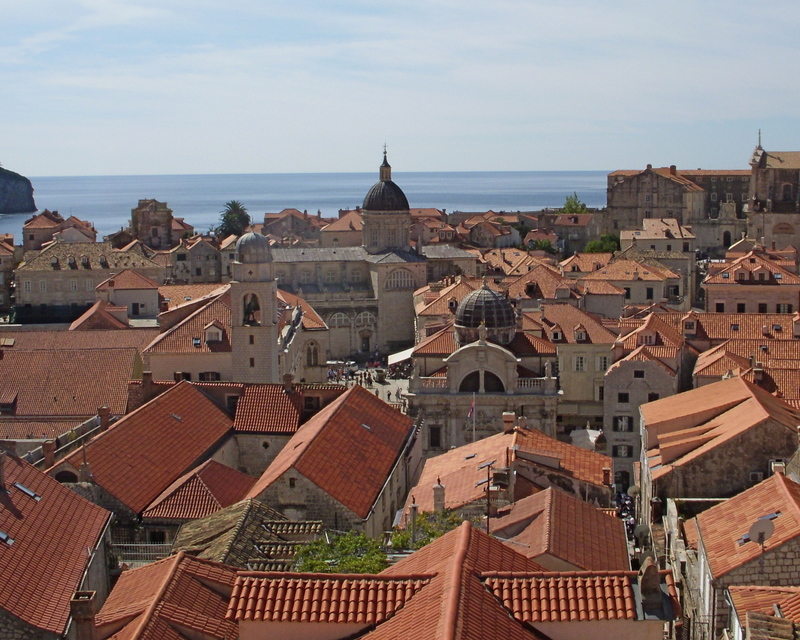}
		\includegraphics[height=50mm]{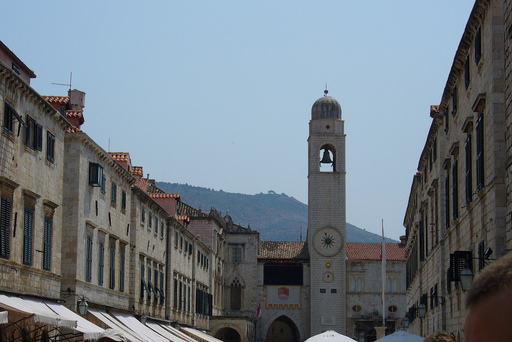}
		}
        
    \vspace{0.5mm}
	\resizebox{\linewidth}{!}{
		\includegraphics[height=50mm]{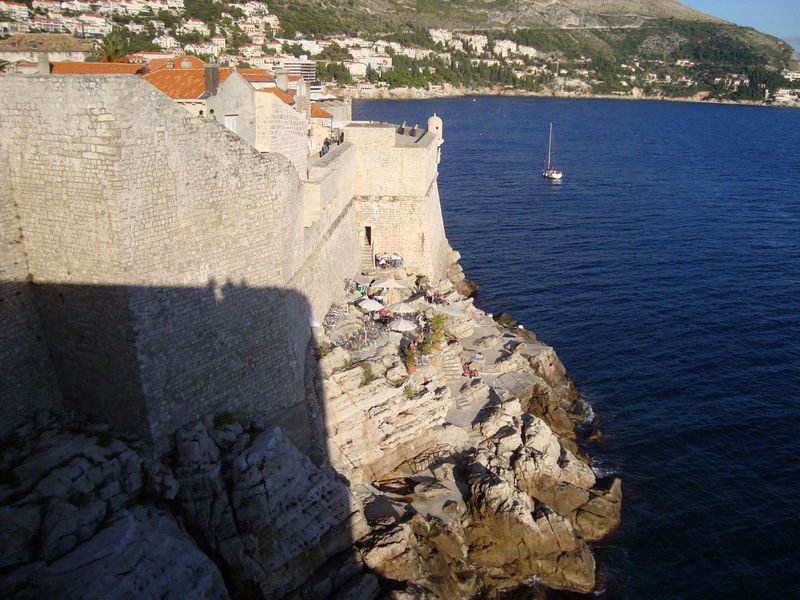}
		\includegraphics[height=50mm]{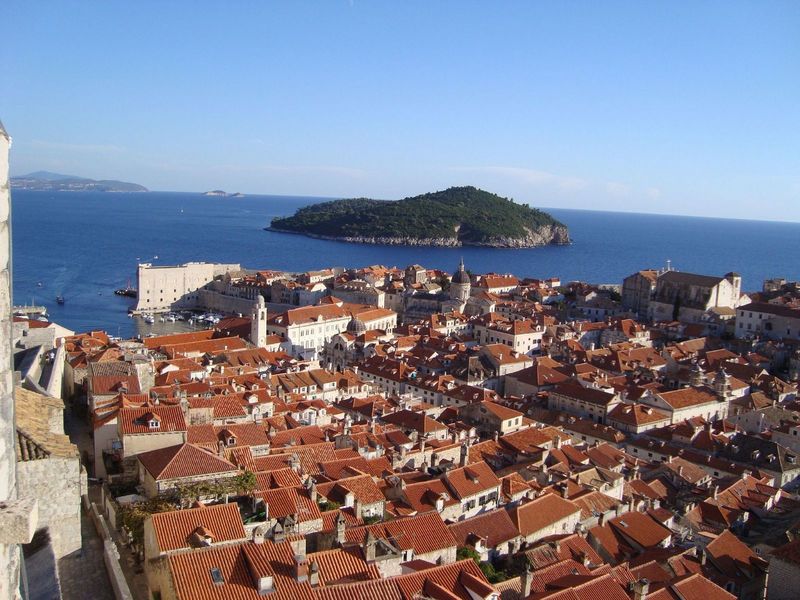}
		\includegraphics[height=50mm]{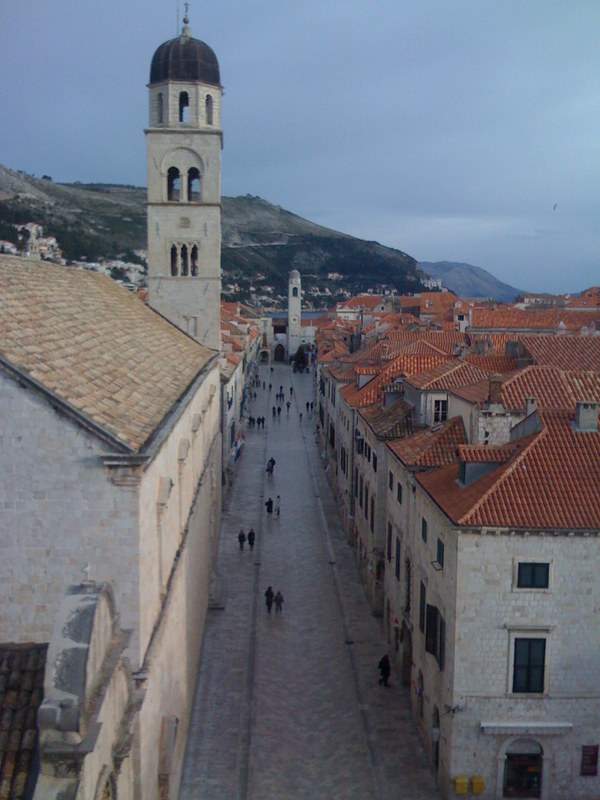}
		\includegraphics[height=50mm]{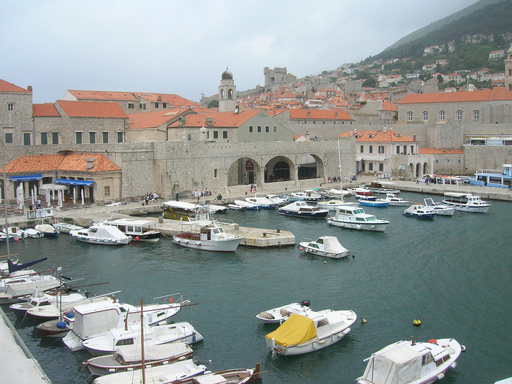}
		\includegraphics[height=50mm]{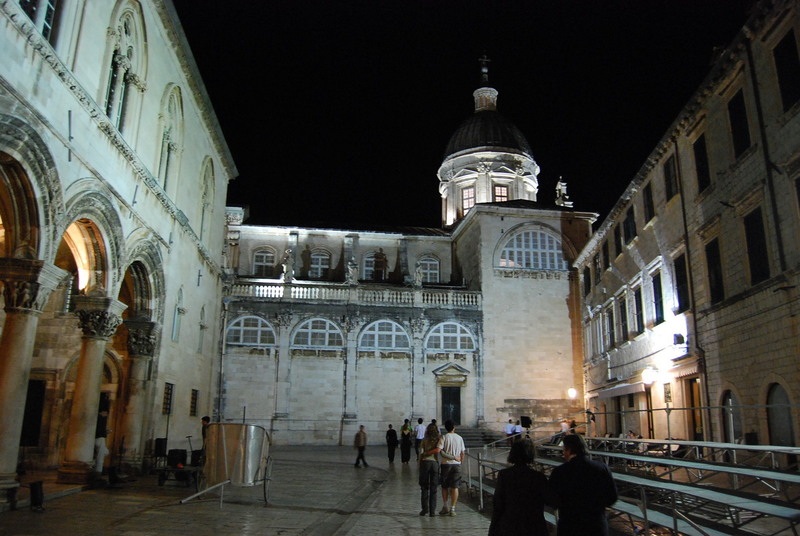}
		}
	\caption{Dubrovnik 6K Dataset - 6,000 images from a variety of camera types in Dubrovnik, Croatia \cite{li2010location}.}
	~
	~
	\end{subfigure}
	
\caption{\textbf{Example images randomly chosen from each dataset.} This illustrates the wide variety of settings and scales and the challenging array of environmental factors such as lighting, occlusion, dynamic objects and weather which are captured in each dataset.}
\label{fig:datasets}
\end{figure*}
	
\begin{table*}[t]
	\centering
\resizebox{\linewidth}{!}{
	\begin{tabular}{l|c|c|c|c|c|c|c|c}
		Dataset & Type & Scale & Imagery & Scenes & Train Images & Test Images & 3-D Points & Spatial Area \\ \hline \hline
		7 Scenes \cite{shotton2013scene} & Indoor & Room & RGB-D sensor (Kinect) & 7 & 26,000 & 17,000 & - & 4$\times$3m \\
		Cambridge Landmarks \cite{kendall2015posenet} & Outdoor & Street & Mobile phone camera & 6 & 8,380 & 4,841 & 2,097,191 & 100$\times$500m \\
		Dubrovnik 6K \cite{li2012worldwide} & Outdoor & Small town & Internet images (Flikr) & 1 & 6,044 & 800 & 2,106,456 & 1.5$\times$1.5km \\
	\end{tabular}}
	\caption{\textbf{Summary of the localisation datasets used in this paper's experiments.} These datasets are all publicly available. They demonstrate our method's performance over a range of scales for both indoor and outdoor applications.}
	\label{tbl:datasets}
\end{table*}

\begin{table*}[t]
	\centering
	\resizebox{\textwidth}{!}{
	\begin{tabular}{l|ccc|ccc}
\multicolumn{1}{c|}{} & \multicolumn{3}{c|}{\textit{Cambridge Landmarks, King's College} \cite{kendall2015posenet}} & \multicolumn{3}{c}{\textit{Dubrovnik 6K} \cite{li2010location}}\\
& \multicolumn{2}{c}{Median Error} & Accuracy & \multicolumn{2}{c}{Median Error} & Accuracy \\
Loss function & x [$m$] & q [$\degree$] & $<2m, 5\degree$ [$\%$] & x [$m$] & q [$\degree$] & $<10m, 10\degree$ [$\%$] \\ \hline \hline
 Linear sum, $\beta=500$ \eqn{loss1} & 1.52 & 1.19 & 65.0\% & 13.1 & 4.68 & 30.1\% \\
 Learn weighting with homoscedastic uncertainty \eqn{loss4} & 0.99 & 1.06 & 85.3\% & 9.88 & 4.73 & 41.7\% \\ \hline
 Reprojection loss & \multicolumn{6}{c}{does not converge} \\ \hline
 Learn weighting pretraining $\mapsto$ Reprojection loss \eqn{loss_reproject} & 0.88 & 1.04 & 90.3\% & 7.90 & 4.40 & 48.6\% \\
	\end{tabular}}
	\caption{\textbf{Comparison of different loss functions.} We use an $L_1$ distance for the residuals in each loss. \textit{Linear sum} combines position and orientation losses with a constant scaling parameter $\beta$ \cite{kendall2015modelling} and is defined in \eqn{loss1}. Learn weighting is the loss function in \eqn{loss4} which learns to combine position and orientation using homoscedastic uncertainty. Reprojection error implicitly combines rotation and translation by using the reprojection error of the scene geometry as the loss \eqn{loss_reproject}. We find that homoscedastic uncertainty is able to learn an effective weighting between position and orientation quantities. The reprojection loss was not able to converge from random initialisation. However, when used to fine-tune a network pretrained with \eqn{loss4} it yields the best results.}
	\label{tbl:losses}
\end{table*}

\subsubsection{Regression norm}
\label{sec:norm}

An important choice for these losses is the regression norm, $\left\lVert~\right\rVert_\gamma$. Typically, deep learning models use an $L_1 = \left\lVert~\right\rVert_1$ or $L_2 = \left\lVert~\right\rVert_2$. We can also use robust norms such as Huber's loss \cite{huber2011robust} and Tukey's loss \cite{hoaglin1983understanding}, which have been successfully applied to deep learning \cite{belagiannis2015robust}. For camera pose regression, we found that they negatively impacted performance by over-attenuating difficult examples. We suspect that for more noisy datasets these robust regression functions might be beneficial. With the datasets used in this paper, we found the $L_1$ norm to perform best and therefore use use $\gamma=1$. It does not increase quadratically with magnitude or over-attenuate large residuals.

\section{Experiments}
\label{sec:exp}

To train and benchmark our model on a number of datasets we rescale the input images such that the shortest side is of length 256. We normalise the images so that input pixel intensities range from $-1$ to $1$. We train our PoseNet architecture using an implementation in TensorFlow \cite{tensorflow2015-whitepaper}. All models are optimised end-to-end with ADAM \cite{kingma2014adam} using the default parameters and a learning rate of $1 \times 10^{-4}$. We train each model until the training loss converges. We use a batch size of 64 on a NVIDIA Titan X (Pascal) GPU, training takes approximately 20k - 100k iterations, or 4 hours - 1 day.

\subsection{Datasets}
\label{sec:data}

Deep learning performs extremely well on large datasets. However annotating ground truth labels on these datasets is often expensive or very labour intensive. We can leverage structure from motion \cite{Snavely08IJCV}, or similar algorithms \cite{shotton2013scene}, to autonomously generate training labels (camera poses) from image data \cite{kendall2015posenet}. We use three datasets to benchmark our approach. These datasets are summarised in \tbl{datasets} and example imagery is shown in \fig{datasets}. We use these datasets to demonstrate our method's performance across a range of settings and scales. We endeavour to demonstrate the general applicability of the approach.

\textbf{Cambridge Landmarks} \cite{kendall2015posenet} provides labelled video data to train and test pose regression algorithms in an outdoor urban setting. It was collected using a smart phone and structure from motion was used to generate the pose labels \cite{wu2013towards}. Significant urban clutter such as pedestrians and vehicles were present and data was collected from many different points in time representing different lighting and weather conditions. Train and test images are taken from distinct walking paths and not sampled from the same trajectory making the regression challenging.

\textbf{7 Scenes} \cite{shotton2013scene} is an indoor dataset which was collected with a Kinect RGB-D sensor. Ground truth poses were computed using Kinect Fusion \cite{shotton2013scene}. The dataset contains seven scenes which were captured around an office building. Each scene typically consists of a single room. The dataset was originally created for RGB-D relocalization. It is extremely challenging for purely visual relocalization using SIFT-like features, as it contains many ambiguous textureless features.

\textbf{Dubrovnik 6K} \cite{li2012worldwide} is a dataset consisting of 6,044 train and 800 test images which were obtained from the internet. They are taken from Dubrovnik's old town in Croatia which is a UNESCO world heritage site. The images are predominantly captured by tourists with a wide variety of camera types. Ground truth poses for this dataset were computed using structure from motion.

\begin{table*}[t]
\begin{center}
\resizebox{\linewidth}{!}{
\begin{tabular}{l|c|c c c c c c}
 & Area or & Active Search & PoseNet & Bayesian & PoseNet & PoseNet (this work) & PoseNet (this work)\\
Scene & Volume & (SIFT) \cite{sattler2016efficient} & ($\beta$ weight) \cite{kendall2015posenet} & PoseNet \cite{kendall2015modelling} &  Spatial LSTM \cite{walch2016image} & Learn $\sigma^2$ Weight & Geometric Reprojection\\
\hline \hline
Great Court	 		& $8000m^2$		& -- 				 &	--				  &	--			   	   & -- 				& 7.00m, 3.65\degree & 6.83m, 3.47\degree \\
King's College 		& $5600m^2$ 	& 0.42m, 0.55\degree & 1.66m, 4.86\degree & 1.74m, 4.06\degree & 0.99m, 3.65\degree & 0.99m, 1.06\degree & 0.88m, 1.04\degree \\
Old Hospital 		& $2000m^2$ 	& 0.44m, 1.01\degree & 2.62m, 4.90\degree & 2.57m, 5.14\degree & 1.51m, 4.29\degree & 2.17m, 2.94\degree & 3.20m, 3.29\degree \\
Shop Fa\c cade 		& $875m^2$ 		& 0.12m, 0.40\degree & 1.41m, 7.18\degree & 1.25m, 7.54\degree & 1.18m, 7.44\degree & 1.05m, 3.97\degree & 0.88m, 3.78\degree \\
St Mary's Church 	& $4800m^2$ 	& 0.19m, 0.54\degree & 2.45m, 7.96\degree & 2.11m, 8.38\degree & 1.52m, 6.68\degree & 1.49m, 3.43\degree & 1.57m, 3.32\degree \\
Street 				& $50000m^2$ 	& 0.85m, 0.83\degree & -- 				  & -- 				   & --					& 20.7m, 25.7\degree & 20.3m, 25.5\degree \\
\hline
\multicolumn{5}{c}{}\\
\hline
Chess 		& $6m^3$ 	& 0.04m, 1.96\degree & 0.32m, 6.60\degree & 0.37m, 7.24\degree & 0.24m, 5.77\degree & 0.14m, 4.50\degree & 0.13m, 4.48\degree \\
Fire 		& $2.5m^3$ 	& 0.03m, 1.53\degree & 0.47m, 14.0\degree & 0.43m, 13.7\degree & 0.34m, 11.9\degree & 0.27m, 11.8\degree & 0.27m, 11.3\degree \\
Heads 		& $1m^3$ 	& 0.02m, 1.45\degree & 0.30m, 12.2\degree & 0.31m, 12.0\degree & 0.21m, 13.7\degree & 0.18m, 12.1\degree & 0.17m, 13.0\degree \\
Office 		& $7.5m^3$ 	& 0.09m, 3.61\degree & 0.48m, 7.24\degree & 0.48m, 8.04\degree & 0.30m, 8.08\degree & 0.20m, 5.77\degree & 0.19m, 5.55\degree \\
Pumpkin 	& $5m^3$ 	& 0.08m, 3.10\degree & 0.49m, 8.12\degree & 0.61m, 7.08\degree & 0.33m, 7.00\degree & 0.25m, 4.82\degree & 0.26m, 4.75\degree \\
Red Kitchen & $18m^3$	& 0.07m, 3.37\degree & 0.58m, 8.34\degree & 0.58m, 7.54\degree & 0.37m, 8.83\degree & 0.24m, 5.52\degree & 0.23m, 5.35\degree \\
Stairs 		& $7.5m^3$ 	& 0.03m, 2.22\degree & 0.48m, 13.1\degree & 0.48m, 13.1\degree & 0.40m, 13.7\degree & 0.37m, 10.6\degree & 0.35m, 12.4\degree \\
\hline
\end{tabular}}
\end{center}

	\caption{\textbf{Median localization results for the \textit{Cambridge Landmarks} \cite{kendall2015posenet} and \textit{7 Scenes} \cite{shotton2013scene} datasets.} We compare the performance of various RGB-only algorithms. Active search \cite{sattler2016efficient} is a state-of-the-art traditional SIFT keypoint based baseline. We demonstrate a notable improvement over PoseNet's \cite{kendall2015posenet} baseline performance using the learned $\sigma^2$ and reprojection error proposed in this paper, narrowing the margin to the state of the art SIFT technique.}
	\label{tbl:mainresults}
\end{table*}

\subsection{Comparison of loss functions}
\label{sec:loss_exp}

In \tbl{losses} we compare different combinations of losses and regression norms. We compare results for a scene in the Cambridge Landmarks dataset \cite{kendall2015posenet} and the Dubrovnik 6K dataset \cite{li2012worldwide}, which has imagery from a range of cameras.

We find that modelling homoscedastic uncertainty with the loss in \eqn{loss4} is able to effectively learn a weighting between position and orientation. It outperforms the constant weighting used in loss \eqn{loss1}. The reprojection loss in \eqn{loss_reproject} is unable to train the model from a random initialisation. We observe that the model gets stuck in a poor local minima, when using any of the regression norms. However, the reprojection loss is able to improve localisation performance when using weights pretrained with any of the other losses. For example, we can take the best performing model using the loss from \eqn{loss4} and fine tune with the reprojection loss \eqn{loss_reproject}. We observe that this loss is then able to converge effectively. This shows that the reprojection loss is not robust to large residuals. This is because reprojected points can be easily placed far from the image centre if the network makes a poor pose prediction. Therefore, we recommend the following two-step training scheme:
\begin{enumerate}[topsep=1pt,itemsep=0ex,partopsep=1ex,parsep=1ex]
  \item Train the model using the loss in \eqn{loss4}, learning the weighting between position and orientation.
  \item If the scene geometry is known (for example from structure from motion or RGBD camera data) then fine-tune the model using the reprojection loss in \eqn{loss_reproject}.
\end{enumerate}

\subsection{Benchmarking localisation accuracy}
\label{sec:benchmark}

In \tbl{mainresults} we show that our geometry based loss outperforms the original PoseNet's naive loss function \cite{kendall2015posenet}. We observe a consistent and significant improvement across both indoor \textit{7 Scenes} outdoor \textit{Cambridge Landmarks} datasets. We conclude that we can simultaneously learn both position and orientation more effectively by considering scene geometry. The improvement is notably more pronounced for the 7Scenes dataset. We believe this is due to the significantly larger amount of training data for each scene in this dataset, compared with Cambridge Landmarks. We also outperform the improved PoseNet architecture with spatial LSTMs \cite{walch2016image}. However, this method is complimentary to the loss functions in this paper, and it would be interesting to explore the union of these ideas.

We observe a difference in relative performance between position and orientation when optimising with respect to reprojection error \eqn{loss_reproject} or homoscedastic uncertainty \eqn{loss4}. Overall, optimising reprojection loss improves rotation accuracy, sometimes at the expense of some positional precision.

\subsection{Comparison to SIFT-feature approaches}

\tbl{mainresults} also compares to a state-of-the-art traditional SIFT feature based localisation algorithm, Active Search \cite{sattler2016efficient}. This method outperforms PoseNet, and is effective in feature-rich outdoor environments. However, in the 7Scenes dataset this deficit is less pronounced. The indoor scenes contain much fewer point features and there is significantly more training data. As an explanation for the deficit in these results, PoseNet only uses $256\times 256$ pixel images, while SIFT based methods require images of a few mega-pixels in size \cite{sattler2016efficient}. Additionally, PoseNet is able to localise an image in $5ms$, scaling constantly with scene area, while traditional SIFT feature approaches require over $100ms$, and scale with scene size \cite{sattler2016efficient}.

In \tbl{dubrovnik_results} we compare our approach on the Dubrovnik dataset to other geometric techniques which localise by registering SIFT features \cite{lowe2004distinctive} to a large 3-D model \cite{li2012worldwide}. Although our method improves significantly over the original PoseNet model, it is still yet to reach the fine grained accuracy of these methods \cite{svarm2014accurate,zeisl2015camera,sattler2011fast,li2010location}. We hypothesise that this is due to a lack of training data, with only 6k images across the town. However, our algorithm is significantly faster than these approaches. Furthermore, it is worth noting that PoseNet only sees a $256\times256$ resolution image, while these methods register the full size images, often with a few million pixels.
	
\begin{table}[t]
\resizebox{\linewidth}{!}{
	\centering
	\begin{tabular}{l|cc|cc}
		& \multicolumn{2}{c|}{Position} & \multicolumn{2}{c}{Orientation} \\
		Method & Mean [m] & Median [m] & Mean [$\degree$] & Median [$\degree$] \\ \hline \hline
		PoseNet (this work) & 40.0 & 7.9 & 11.2 & 4.4 \\
		APE \cite{svarm2014accurate} & - & 0.56 & - & - \\
		Voting \cite{zeisl2015camera} & - & 1.69 & - & - \\
		Sattler, et al. \cite{sattler2011fast} & 14.9 & 1.3 & - & - \\
		P2F \cite{li2010location} & 18.3 & 9.3 & - & - \\
	\end{tabular}}
	\caption{\textbf{Localisation results on the Dubrovnik dataset} \cite{li2012worldwide}, comparing to a number of state-of-the-art point-feature techniques. Our method is the first deep learning approach to benchmark on this challenging dataset. We achieve comparable performance, while our method only requires a 256$\times$256 pixel image and is much faster to compute.}
	\label{tbl:dubrovnik_results}
\end{table}

\section{Conclusions}

We have investigated a number of loss functions for learning to regress position and orientation simultaneously with scene geometry. We present an algorithm for training PoseNet which does not require any hyper-parameter tuning. We demonstrate PoseNet's efficacy on three large scale datasets. We observe a large improvement of results compared to the original loss proposed by PoseNet, narrowing the performance gap to traditional point-feature approaches. 

For many applications which require localization, such as mobile robotics, video data is readily available. Ultimately, we would like to extend the architecture to video input with further use of multi-view stereo \cite{hartley2003multiple}.

{\footnotesize
\bibliographystyle{ieee}
\bibliography{bib}

\begin{thebibliography}{10}\itemsep=-1pt

\bibitem{tensorflow2015-whitepaper}
M.~Abadi, A.~Agarwal, P.~Barham, E.~Brevdo, Z.~Chen, C.~Citro, G.~S. Corrado,
  A.~Davis, J.~Dean, M.~Devin, S.~Ghemawat, I.~Goodfellow, A.~Harp, G.~Irving,
  M.~Isard, Y.~Jia, R.~Jozefowicz, L.~Kaiser, M.~Kudlur, J.~Levenberg,
  D.~Man\'{e}, R.~Monga, S.~Moore, D.~Murray, C.~Olah, M.~Schuster, J.~Shlens,
  B.~Steiner, I.~Sutskever, K.~Talwar, P.~Tucker, V.~Vanhoucke, V.~Vasudevan,
  F.~Vi\'{e}gas, O.~Vinyals, P.~Warden, M.~Wattenberg, M.~Wicke, Y.~Yu, and
  X.~Zheng.
\newblock {TensorFlow}: Large-scale machine learning on heterogeneous systems,
  2015.
\newblock Software available from tensorflow.org.

\bibitem{altmann2005rotations}
S.~L. Altmann.
\newblock {\em Rotations, quaternions, and double groups}.
\newblock Courier Corporation, 2005.

\bibitem{BL15}
A.~Babenko and V.~Lempitsky.
\newblock Aggregating deep convolutional features for image retrieval.
\newblock In {\em International Conference on Computer Vision (ICCV)}, 2015.

\bibitem{BSCL14}
A.~Babenko, A.~Slesarev, A.~Chigorin, and V.~Lempitsky.
\newblock Neural codes for image retrieval.
\newblock In {\em European Conference on Computer Vision}, 2014.

\bibitem{belagiannis2015robust}
V.~Belagiannis, C.~Rupprecht, G.~Carneiro, and N.~Navab.
\newblock Robust optimization for deep regression.
\newblock In {\em International Conference on Computer Vision (ICCV)}, pages
  2830--2838. IEEE, 2015.

\bibitem{ChenBaatz11CVPR}
D.~Chen, G.~Baatz, K.~K\"oser, S.~Tsai, R.~Vedantham, T.~Pylv\"an\"ainen,
  K.~Roimela, X.~Chen, J.~Bach, M.~Pollefeys, B.~Girod, and R.~Grzeszczuk.
\newblock City-scale landmark identification on mobile devices.
\newblock In {\em Proceedings of the IEEE Conference on Computer Vision and
  Pattern Recognition}, 2011.

\bibitem{Choudhary12ECCV}
S.~Choudhary and P.~J. Narayanan.
\newblock Visibility probability structure from sfm datasets and applications.
\newblock In {\em European Conference on Computer Vision}, 2012.

\bibitem{clark2017vidloc}
R.~Clark, S.~Wang, A.~Markham, N.~Trigoni, and H.~Wen.
\newblock Vidloc: 6-dof video-clip relocalization.
\newblock {\em arXiv preprint arXiv:1702.06521}, 2017.

\bibitem{cummins2008fab}
M.~Cummins and P.~Newman.
\newblock {FAB-MAP}: Probabilistic localization and mapping in the space of
  appearance.
\newblock {\em The International Journal of Robotics Research}, 27(6):647--665,
  2008.

\bibitem{Delhumeau-ACMMM13}
J.~Delhumeau, P.-H. Gosselin, H.~J{\'e}gou, and P.~P{\'e}rez.
\newblock {Revisiting the VLAD image representation}.
\newblock In {\em {ACM Multimedia}}, Barcelona, Spain, Oct. 2013.

\bibitem{deng2009imagenet}
J.~Deng, W.~Dong, R.~Socher, L.-J. Li, K.~Li, and L.~Fei-Fei.
\newblock Imagenet: A large-scale hierarchical image database.
\newblock In {\em Proceedings of the IEEE Conference on Computer Vision and
  Pattern Recognition}, pages 248--255. IEEE, 2009.

\bibitem{engel2014lsd}
J.~Engel, T.~Sch{\"o}ps, and D.~Cremers.
\newblock {LSD-SLAM}: Large-scale direct monocular slam.
\newblock In {\em European Conference on Computer Vision}, pages 834--849.
  Springer, 2014.

\bibitem{GWGL14}
Y.~Gong, L.~Wang, R.~Guo, and S.~Lazebnik.
\newblock Multi-scale orderless pooling of deep convolutional activation
  features.
\newblock In {\em European Conference on Computer Vision}, 2014.

\bibitem{hartley2003multiple}
R.~Hartley and A.~Zisserman.
\newblock {\em Multiple view geometry in computer vision}.
\newblock Cambridge university press, 2003.

\bibitem{hoaglin1983understanding}
D.~C. Hoaglin, F.~Mosteller, and J.~W. Tukey.
\newblock {\em Understanding robust and exploratory data analysis}, volume~3.
\newblock Wiley New York, 1983.

\bibitem{huber2011robust}
P.~J. Huber.
\newblock {\em Robust statistics}.
\newblock Springer, 2011.

\bibitem{Jegou-CVPR10}
H.~J\'egou, M.~Douze, C.~Schmid, and P.~P\'erez.
\newblock Aggregating local descriptors into a compact image representation.
\newblock In {\em Proceedings of the IEEE Conference on Computer Vision and
  Pattern Recognition}, pages 3304--3311, jun 2010.

\bibitem{Jegou-PAMI12}
H.~Jegou, F.~Perronnin, M.~Douze, J.~S{\'a}nchez, P.~Perez, and C.~Schmid.
\newblock Aggregating local image descriptors into compact codes.
\newblock {\em IEEE transactions on pattern analysis and machine intelligence},
  34(9):1704--1716, 2012.

\bibitem{kendall2015modelling}
A.~Kendall and R.~Cipolla.
\newblock Modelling uncertainty in deep learning for camera relocalization.
\newblock {\em arXiv preprint arXiv:1509.05909}, 2015.

\bibitem{kendall2017uncertainties}
A.~Kendall and Y.~Gal.
\newblock What uncertainties do we need in bayesian deep learning for computer
  vision?
\newblock {\em arXiv preprint arXiv:1703.04977}, 2017.

\bibitem{kendall2017multi}
A.~Kendall, Y.~Gal, and R.~Cipolla.
\newblock Multi-task learning using uncertainty to weigh losses for scene
  geometry and semantics.
\newblock {\em arXiv preprint arXiv:1705.07115}, 2017.

\bibitem{kendall2015posenet}
A.~Kendall, M.~Grimes, and R.~Cipolla.
\newblock Posenet: A convolutional network for real-time 6-dof camera
  relocalization.
\newblock {\em arXiv preprint arXiv:1505.07427}, 2015.

\bibitem{kingma2014adam}
D.~Kingma and J.~Ba.
\newblock Adam: A method for stochastic optimization.
\newblock {\em arXiv preprint arXiv:1412.6980}, 2014.

\bibitem{klein2007parallel}
G.~Klein and D.~Murray.
\newblock Parallel tracking and mapping for small ar workspaces.
\newblock In {\em Mixed and Augmented Reality, IEEE and ACM International
  Symposium on}, pages 225--234. IEEE, 2007.

\bibitem{li2017indoor}
R.~Li, Q.~Liu, J.~Gui, D.~Gu, and H.~Hu.
\newblock Indoor relocalization in challenging environments with dual-stream
  convolutional neural networks.
\newblock {\em IEEE Transactions on Automation Science and Engineering}, 2017.

\bibitem{li2012worldwide}
Y.~Li, N.~Snavely, D.~Huttenlocher, and P.~Fua.
\newblock Worldwide pose estimation using 3d point clouds.
\newblock In {\em European Conference on Computer Vision}, pages 15--29.
  Springer, 2012.

\bibitem{Li12ECCV}
Y.~Li, N.~Snavely, D.~Huttenlocher, and P.~Fua.
\newblock {Worldwide Pose Estimation Using 3D Point Clouds}.
\newblock In {\em European Conference on Computer Vision}, 2012.

\bibitem{Li10ECCV}
Y.~Li, N.~Snavely, and D.~P. Huttenlocher.
\newblock {Location Recognition using Prioritized Feature Matching}.
\newblock In {\em European Conference on Computer Vision}, 2010.

\bibitem{li2010location}
Y.~Li, N.~Snavely, and D.~P. Huttenlocher.
\newblock Location recognition using prioritized feature matching.
\newblock In {\em European Conference on Computer Vision}, pages 791--804.
  Springer, 2010.

\bibitem{lowe2004distinctive}
D.~G. Lowe.
\newblock Distinctive image features from scale-invariant keypoints.
\newblock {\em International journal of computer vision}, 60(2):91--110, 2004.

\bibitem{melekhov2017relative}
I.~Melekhov, J.~Kannala, and E.~Rahtu.
\newblock Relative camera pose estimation using convolutional neural networks.
\newblock {\em arXiv preprint arXiv:1702.01381}, 2017.

\bibitem{moser2008place}
E.~I. Moser, E.~Kropff, and M.-B. Moser.
\newblock Place cells, grid cells, and the brain's spatial representation
  system.
\newblock {\em Annu. Rev. Neurosci.}, 31:69--89, 2008.

\bibitem{mur2015orb}
R.~Mur-Artal, J.~Montiel, and J.~D. Tard{\'o}s.
\newblock Orb-slam: a versatile and accurate monocular slam system.
\newblock {\em IEEE Transactions on Robotics}, 31(5):1147--1163, 2015.

\bibitem{newcombe2011dtam}
R.~A. Newcombe, S.~J. Lovegrove, and A.~J. Davison.
\newblock {DTAM}: Dense tracking and mapping in real-time.
\newblock In {\em International Conference on Computer Vision (ICCV)}, pages
  2320--2327. IEEE, 2011.

\bibitem{Nister06CVPR}
D.~Nister and H.~Stewenius.
\newblock Scalable recognition with a vocabulary tree.
\newblock In {\em Proceedings of the IEEE Conference on Computer Vision and
  Pattern Recognition}, 2006.

\bibitem{o1978hippocampus}
J.~O'keefe and L.~Nadel.
\newblock {\em The hippocampus as a cognitive map}, volume~3.
\newblock Clarendon Press Oxford, 1978.

\bibitem{oquab2014learning}
M.~Oquab, L.~Bottou, I.~Laptev, and J.~Sivic.
\newblock Learning and transferring mid-level image representations using
  convolutional neural networks.
\newblock In {\em Proceedings of the IEEE Conference on Computer Vision and
  Pattern Recognition}, pages 1717--1724. IEEE, 2014.

\bibitem{Philbin07CVPR}
J.~Philbin, O.~Chum, M.~Isard, J.~Sivic, and A.~Zisserman.
\newblock {Object Retrieval with Large Vocabularies and Fast Spatial Matching}.
\newblock In {\em Proceedings of the IEEE Conference on Computer Vision and
  Pattern Recognition}, 2007.

\bibitem{RSMC14}
A.~S. Razavian, J.~Sullivan, A.~Maki, and S.~Carlsson.
\newblock A baseline for visual instance retrieval with deep convolutional
  networks.
\newblock In {\em arXiv:1412.6574}, 2014.

\bibitem{rublee2011orb}
E.~Rublee, V.~Rabaud, K.~Konolige, and G.~Bradski.
\newblock Orb: An efficient alternative to sift or surf.
\newblock In {\em International conference on computer vision (ICCV)}, pages
  2564--2571. IEEE, 2011.

\bibitem{sattler2011fast}
T.~Sattler, B.~Leibe, and L.~Kobbelt.
\newblock Fast image-based localization using direct 2d-to-3d matching.
\newblock In {\em International Conference on Computer Vision}, pages 667--674.
  IEEE, 2011.

\bibitem{Sattler12ECCV}
T.~Sattler, B.~Leibe, and L.~Kobbelt.
\newblock {Improving Image-Based Localization by Active Correspondence Search}.
\newblock 2012.

\bibitem{sattler2016efficient}
T.~Sattler, B.~Leibe, and L.~Kobbelt.
\newblock Efficient \& effective prioritized matching for large-scale
  image-based localization.
\newblock {\em IEEE Transactions on Pattern Analysis and Machine Intelligence},
  2016.

\bibitem{Sattler14ECCV}
T.~Sattler, C.~Sweeney, and M.~Pollefeys.
\newblock On sampling focal length values to solve the absolute pose problem.
\newblock In {\em European Conference on Computer Vision}, pages 828--843.
  Springer, 2014.

\bibitem{Schindler07CVPR}
G.~Schindler, M.~Brown, and R.~Szeliski.
\newblock City-scale location recognition.
\newblock In {\em Computer Vision and Pattern Recognition, 2007. CVPR'07. IEEE
  Conference on}, pages 1--7. IEEE, 2007.

\bibitem{shotton2013scene}
J.~Shotton, B.~Glocker, C.~Zach, S.~Izadi, A.~Criminisi, and A.~Fitzgibbon.
\newblock Scene coordinate regression forests for camera relocalization in
  {RGB-D} images.
\newblock In {\em Computer Vision and Pattern Recognition (CVPR), IEEE
  Conference on}, pages 2930--2937. IEEE, 2013.

\bibitem{Sivic03ICCV}
J.~Sivic, A.~Zisserman, et~al.
\newblock Video google: A text retrieval approach to object matching in videos.
\newblock In {\em International Conference on Computer Vision (ICCV)},
  volume~2, pages 1470--1477, 2003.

\bibitem{Snavely08IJCV}
N.~Snavely, S.~M. Seitz, and R.~Szeliski.
\newblock Modeling the world from internet photo collections.
\newblock {\em International Journal of Computer Vision}, 80(2):189--210, 2008.

\bibitem{Svarm14CVPR}
L.~Svarm, O.~Enqvist, M.~Oskarsson, and F.~Kahl.
\newblock Accurate localization and pose estimation for large 3d models.
\newblock In {\em Proceedings of the IEEE Conference on Computer Vision and
  Pattern Recognition}, pages 532--539, 2014.

\bibitem{svarm2014accurate}
L.~Svarm, O.~Enqvist, M.~Oskarsson, and F.~Kahl.
\newblock Accurate localization and pose estimation for large 3d models.
\newblock In {\em Proceedings of the IEEE Conference on Computer Vision and
  Pattern Recognition}, pages 532--539, 2014.

\bibitem{szegedy2014going}
C.~Szegedy, W.~Liu, Y.~Jia, P.~Sermanet, S.~Reed, D.~Anguelov, D.~Erhan,
  V.~Vanhoucke, and A.~Rabinovich.
\newblock Going deeper with convolutions.
\newblock {\em arXiv preprint arXiv:1409.4842}, 2014.

\bibitem{Tolias16ICLR}
G.~Tolias, R.~Sicre, and H.~JÃ©gou.
\newblock Particular object retrieval with integral max-pooling of cnn
  activations.
\newblock In {\em ICLR}, 2016.

\bibitem{Torii13CVPR}
A.~Torii, J.~Sivic, T.~Pajdla, and M.~Okutomi.
\newblock Visual place recognition with repetitive structures.
\newblock In {\em Proceedings of the IEEE conference on computer vision and
  pattern recognition}, pages 883--890, 2013.

\bibitem{walch2016image}
F.~Walch, C.~Hazirbas, L.~Leal-Taix{\'e}, T.~Sattler, S.~Hilsenbeck, and
  D.~Cremers.
\newblock Image-based localization with spatial lstms.
\newblock {\em arXiv preprint arXiv:1611.07890}, 2016.

\bibitem{weyand2016planet}
T.~Weyand, I.~Kostrikov, and J.~Philbin.
\newblock Planet-photo geolocation with convolutional neural networks.
\newblock In {\em European Conference on Computer Vision}, pages 37--55.
  Springer, 2016.

\bibitem{wu2013towards}
C.~Wu.
\newblock Towards linear-time incremental structure from motion.
\newblock In {\em 3D Vision-3DV 2013, 2013 International Conference on}, pages
  127--134. IEEE, 2013.

\bibitem{zeisl2015camera}
B.~Zeisl, T.~Sattler, and M.~Pollefeys.
\newblock Camera pose voting for large-scale image-based localization.
\newblock In {\em International Conference on Computer Vision (ICCV)}, 2015.

\end{thebibliography}
}

\end{document}